%% file: main.tex
\documentclass[10pt,twocolumn,letterpaper]{article}

\usepackage{iccv}              


\definecolor{iccvblue}{rgb}{0.21,0.49,0.74}
\usepackage[pagebackref,breaklinks,colorlinks,allcolors=iccvblue]{hyperref}

\usepackage{xspace}
\usepackage{amssymb}
\usepackage{amsmath}
\usepackage{amsfonts}
\usepackage{wrapfig}
\usepackage{algorithm}
\usepackage{algorithmic}

\usepackage{multicol}
\usepackage{multirow}
\usepackage{makecell}
\usepackage{wrapfig}
\usepackage{tabularx}
\usepackage{adjustbox}

\usepackage{pifont}
\usepackage{color}
\usepackage{colortbl}

\usepackage{caption}

\definecolor{lightgray}{rgb}{0.8, 0.8, 0.8}
\definecolor{lgray}{rgb}{0.66, 0.66, 0.66}

\definecolor{lblu_tab}{RGB}{225, 235, 246}
\definecolor{orange_vitad}{RGB}{222, 131, 68}
\definecolor{blue_vitad}{RGB}{106, 153, 208}
\definecolor{trajectory_green}{RGB}{126, 171, 85}
\definecolor{trajectory_yellow}{RGB}{245, 194, 66}
\definecolor{tab_others}{RGB}{235, 235, 235}
\definecolor{tab_ours}{RGB}{225, 235, 246}

\definecolor{whit_tab}{RGB}{255, 255, 255}
\definecolor{gray_tab}{RGB}{246, 246, 246}
\definecolor{oran_tab}{RGB}{252, 242, 237}
\definecolor{blue_tab}{RGB}{227, 240, 251}

\def\method{PointSeg}

\def\paperID{1} 
\def\confName{ICCV}
\def\confYear{2025}

\title{\method: A Training-Free Paradigm for 3D Scene Segmentation via Foundation Models}

\author{
  Qingdong He$^{1 \dagger}$
  ~~ Jinlong Peng$^1$ 
  ~~ Zhengkai Jiang$^1$
  ~~ Xiaobin Hu$^1$ 
  ~~ Jiangning Zhang$^1$
   \\ \vspace{4pt}
   \normalsize $^1$Tencent Youtu Lab  \\
}

\begin{document}

\maketitle
\renewcommand{\thefootnote}{}
\footnotetext[1]{$\dagger~$Corresponding author.}

\input{secs/0_abstract}    
\input{secs/1_introduction}
\input{secs/2_related_work}
\input{secs/3_method}
\input{secs/4_experiment}
\input{secs/5_conclusion}

{
    \small
    \bibliographystyle{ieeenat_fullname}
    \bibliography{main}
}
\input{secs/X_suppl}

\end{document}

%% file: secs/0_abstract.tex
\begin{abstract}
Recent success of vision foundation models have shown promising performance for the 2D perception tasks. However, it is difficult to train a 3D foundation network directly due to the limited dataset and it remains under explored whether existing foundation models can be lifted to 3D space seamlessly. In this paper, we present PointSeg, a novel training-free paradigm that leverages off-the-shelf vision foundation models to address 3D scene perception tasks. PointSeg can segment anything in 3D scene by acquiring accurate 3D prompts to align their corresponding pixels across frames. Concretely, we design a two-branch prompts learning structure to construct the 3D point-box prompts pairs, combining with the bidirectional matching strategy for accurate point and proposal prompts generation. Then, we perform the iterative post-refinement adaptively when cooperated with different vision foundation models. Moreover, we design a affinity-aware merging algorithm to improve the final ensemble masks. PointSeg demonstrates impressive segmentation performance across various datasets, all without training. Specifically, our approach significantly surpasses the state-of-the-art specialist training-free model by 16.3$\%$, 14.9$\%$, and 15$\%$ mAP on ScanNet, ScanNet++, and KITTI-360 datasets, respectively. On top of that, PointSeg can incorporate with various foundation models and even surpasses the specialist training-based methods by 5.6$\%$-8$\%$ mAP across various datasets, serving as an effective generalist model.
\end{abstract}

%% file: secs/1_introduction.tex
\section{Introduction} \label{sec:Introduction}
3D scene segmentation plays a vital role in many applications, such as autonomous driving, augmented reality and room navigation. To tackle the challenges in 3D scene segmentation, most of the previous methods~\citep{kundu2020virtual,jiang2020pointgroup,rozenberszki2022language,kolodiazhnyi2024top,liang2021instance,schult2023mask3d,sun2023superpoint,vu2022softgroup++} are supervised and heavily rely on precise 3D annotations, which means that they lack the zero-shot capability. Despite recent efforts~\citep{takmaz2023openmask3d,huang2023openins3d,yin2023or,he2024unim} have attempted to explore the zero-shot 3D scene understanding, these approaches either require 3D mask pre-trained networks or domain-specific data training. Consequently, the ability of domain transfer to unfamiliar 3D scenes continues to pose significant challenges. 

\begin{figure}[t]
    \centering
    \includegraphics[width=1\linewidth]{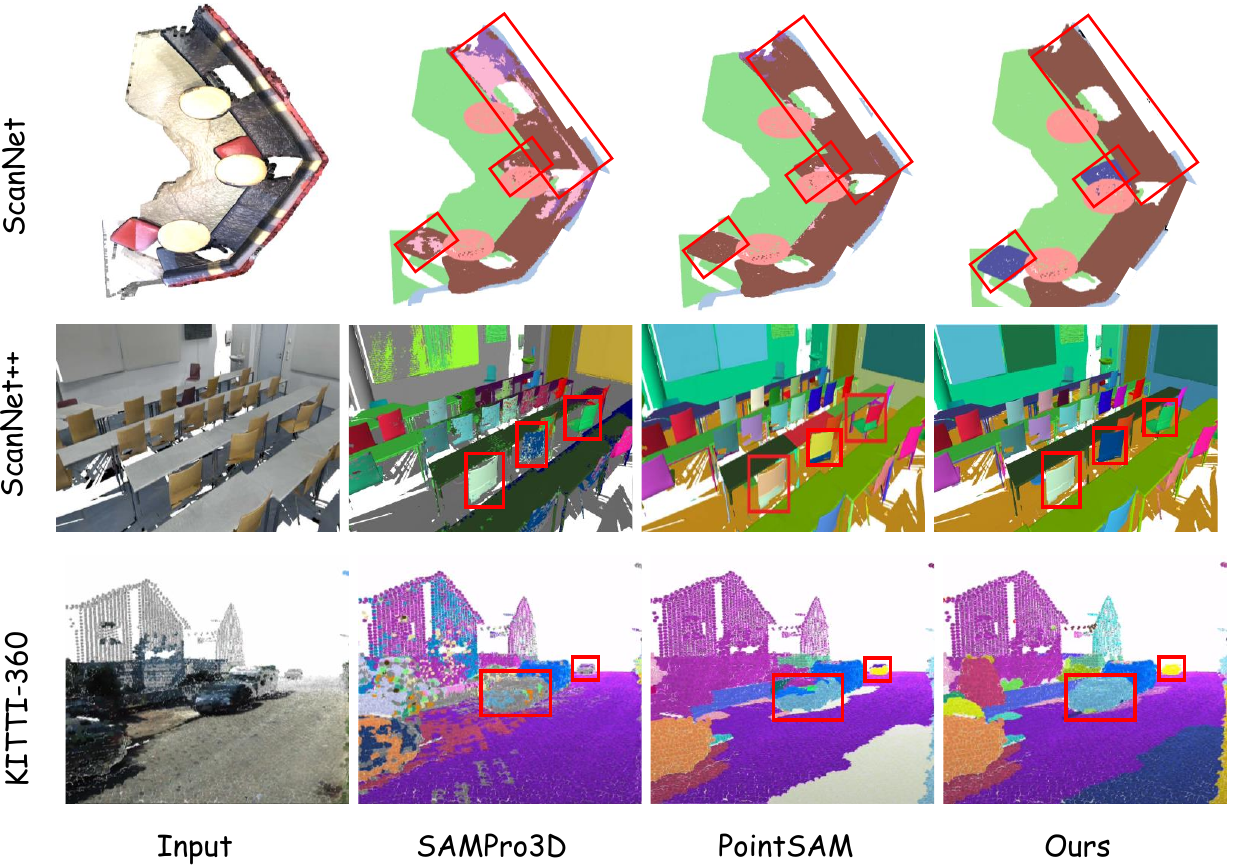}
    \caption{\textbf{Qualitative results comparison} on ScanNet, ScanNet++ and KITTI-360 datasets. Compared to the training-based method PointSAM~\citep{zhou2024point} and the training-free method SAMPro3D~\citep{xu2023sampro3d}, our method can segment objects in 3D scene more completely and accurately.}
    \label{fig::motivation}
\end{figure}

Looking around in the 2D realm, vision foundation models (VFMs)~\citep{radford2021learning,jia2021scaling,oquab2023dinov2,he2022masked} have exploded in growth, attributed to the availability of large-scale datasets and computational resources. And they have demonstrated exceptional generalization capabilities in zero-shot scenarios, along with multifunctional interactivity when combined with human feedback. Most recently,  Segment Anything Model (SAM)~\citep{kirillov2023segment} has managed to attain remarkable performance in class-agnostic segmentation by training on the SA-1B dataset. Then it triggers a series of applications in various tasks and improvements in various aspects~\citep{zhang2023personalize,zou2024segment,xiong2023efficientsam,zhao2023fast,zhang2023faster}. Inspired by this, a natural idea is to also train a foundation model in 3D space. However, this has been hindered by the limited scale of 3D data and the high cost of 3D data collection and annotation~\citep{goyal2021revisiting,chang2015shapenet}. Considering this, we ask: \emph{Is it possible to explore the use of VFMs to effectively tackle a broad spectrum of 3D perception tasks without training, e.g., 3D scene segmentation?}

Following this paradigm, some works have made some early attempts. One line focuses on segmenting 2D frame accurately with different scene deconstruction strategies~\citep{yang2023sam3d,yin2024sai3d,guo2023sam}. Another line tries to learn high-quality 3D points to prompt the SAM by using the projection from 3D to 2D~\citep{xu2023sampro3d}. Though effective, none of these methods essentially acknowledge the facts of 3D scene segmentation in three challenging aspects: (i) 3D prompts are naturally prior to the one in the 2D space, which should be carefully designed rather than a simple projection, (ii) the initial segmentation mask from multiple views might include rough edges and isolated background noises, (iii) local adjacent frames maintain the global consensus, which might be overlooked during the merging process. 

To address these challenges, we present PointSeg, a novel perception framework that effectively incorporates different foundation models for tackling the 3D scene segmentation task without training. The key idea behind PointSeg is to learn accurate 3D point-box prompts pairs to enforce the off-the-shelf foundation models and fully unleash their potential in 3D scene segmentation tasks with three effective components. First, we construct a two-branch prompts learning structure to acquire the 3D point prompts and 3D box prompts respectively. The 3D point prompts are derived from localization abilities of PointLLM~\citep{xu2023pointllm} to provide more explicit prompts in the form of points and 3D box prompts come from the 3D detectors~\citep{shen2023v,wu2023virtual}. Considering that the naive prompts could result in fragmented false-positive masks caused by matching outliers, we propose the bidirectional matching strategy for the generation of accurate point-box prompt pairs. Furthermore, when incorporated with different 2D vision segmentation foundation models, such as SAM 2~\citep{ravi2024sam}, our approach involves the iterative post-refinement to eliminate the coarse boundaries and isolated instances of background noise adaptively. Finally, with the primary aim of segmenting all points within the 3D scene, we employ the affinity-aware merging algorithm to capture pairwise similarity scores based on the 2D masks generated by the vision segmentation foundation models.

Comprehensive experiments on ScanNet~\citep{dai2017scannet}, ScanNet++~\citep{yeshwanth2023scannet++}, and KITTI-360~\citep{liao2022kitti} demonstrate the superior generalization of the proposed PointSeg, surpassing previous specialist training-free model by 14.9$\%$-16.3$\%$ mAP and specialist training-based methods by 5.6$\%$-8$\%$ mAP across different datasets, all without training on domain-specific data. Remarkably, our zero-shot approach yields superior results in comparison to fully-supervised PointSAM~\citep{zhou2024point} trained on synthetic datasets, thereby emphasizing the effectiveness of PointSeg in the segmentation of intricate 3D scene. Moreover, we incorporate different foundation models, \emph{i.e.}, SAM 2~\citep{ravi2024sam}, SAM~\citep{kirillov2023segment}, FastSAM~\citep{zhao2023fast}, MobileSAM~\citep{zhang2023faster}, and EfficientSAM~\citep{xiong2023efficientsam}, into our pipeline, and the performance gain shows that enhancements on 2D images can be seamlessly translated to improve 3D results. We summarize the contributions of our paper as follows:

\begin{itemize}
   \item We present PointSeg, a novel framework for exploring the potential of leveraging various vision foundation models in tackling 3D scene segmentation task, without training or finetuning with 3D data.
   
   \item We design PointSeg as a two-branch prompts learning structure, equipped with three key components, \emph{i.e.}, bidirectional matching based prompts generation, iterative post-refinement and affinity-aware merging, which can effectively unleash the ability of vision foundation
     models to improve the 3D segmentation quality.
     
    \item PointSeg outperforms previous specialist training-based and training-free methods on 3D segmentation task by a large margin, which demonstrates the impressive performance and powerful generalization when incorporated with various foundation models. 
\end{itemize}

%% file: secs/2_related_work.tex
\section{Related Work}
\label{related}
\noindent\textbf{Closed-set 3D Segmentation.} 
Considering the point clouds in 3D space, 3D semantic segmentation task aims to predict a specific category towards the given point~\citep{graham20183d,hu2020randla,kundu2020virtual,li2018pointcnn,qi2017pointnet++,rozenberszki2022language,wang2019dynamic,xu2018spidercnn,wang2015efficient,zhang2023growsp}. 3D instance segmentation task broadens this concept by pinpointing separate entities within the same semantic class and bestowing unique masks upon each object instance~\citep{choy20194d,fan2021scf,han2020occuseg,hou20193d,engelmann20203d,hui2022learning,jiang2020pointgroup,kolodiazhnyi2024top,liang2021instance,schult2023mask3d,sun2023superpoint,vu2022softgroup++,lahoud20193d,yang2019learning}. Among them, Mask3D~\citep{schult2023mask3d} designs a transformer-based network to build the 3D segmentation network and achieves the state-of-the-art performance. Although Mask3D has made significant progress, like previous supervised learning methods, it still necessitates a substantial volume of 3D annotated data for network training. This limitation impedes the method's generalization to open-world scenarios that include new objects from unseen categories. Furthermore, the collection of annotated 3D data is not only costly but sometimes unfeasible due to privacy concerns. Our framework, however, aspires to directly leverage the intrinsic zero-shot potential of SAM for 3D scene segmentation, thereby negating the necessity for further model training.

\noindent\textbf{Open-set 3D Segmentation.} 
Inspired by the success of 2D open-vocabulary segmentation methods~\citep{ghiasi2022scaling,liang2023open}, a series of works~\citep{ding2023pla,huang2023openins3d,takmaz2023openmask3d,peng2023openscene,he2024unim,jiang2022prototypical} have dived to explore the potential of 3D open-vocabulary scene understanding. OpenMask3D~\citep{takmaz2023openmask3d} predicts 3D instance masks with the per-mask feature representations, which can be used for querying instances based on open-vocabulary concepts. OpenIns3D~\citep{huang2023openins3d} employs a Mask-Snap-Lookup scheme to learn class-agnostic mask proposals and generate synthetic scene-level images at multiple scales. On the other hand, the interpolation capabilities of NeRFs~\citep{mildenhall2021nerf} are applied to integrate language with the CLIP feature by LERF~\citep{kerr2023lerf} and DFF~\citep{kobayashi2022decomposing}. And OR-NeRF~\citep{yin2023or} empowers users to segment an object by clicking and subsequently eliminate it from the scene. Although they have achieved encouraging instance segmentation results on indoor scenes with objects similar to the training data, these methods demonstrate a failure in complex scenes with fine-grained objects. In this study, we eliminate the reliance on a pre-trained 3D mask proposal network and instead focus directly on how to leverage the segmentation results of SAM to generate fine-grained 3D masks for 3D scenes.

\noindent\textbf{Segment Anything Model in 3D.} 
The emergence of the Segment Anything Model~\citep{kirillov2023segment,ravi2024sam} have triggered a revolution in the field of 2D segmentation. Having been trained on the extraordinary SA-1B dataset, SAM has garnered a vast amount of knowledge, equipping it to effectively segment unfamiliar images without additional training. Followed by SAM, several works~\citep{zhang2023personalize,zou2024segment,xiong2023efficientsam,zhao2023fast,zhang2023faster,liu2023matcher} have attempted to accelerate or customize the original SAM from different aspects. Recognizing the exceptional capabilities of SAM, various recent research initiatives are working diligently to incorporate SAM into 3D scene segmentation task. Several works~\citep{yang2023sam3d,yin2024sai3d,guo2023sam} attempt to segment each frame individually or constructs a graph based on the superpoints to lift the segmentation results to 3D space.
However, these methods designate pixel prompts that are specific to each frame but do not synchronize across frames. This causes inconsistencies in segmentation across frames and produces substandard 3D segmentation results. Different from these 2D-to-3D lifting methods, SAMPro3D~\citep{xu2023sampro3d} attempts to locate 3D points in scenes as 3D prompts to align their projected pixel prompts across frames. Albeit effective, simply connecting 3D points to 2D space through projection is still too rough for complex scenes. In this paper, we propose a two-branch prompts learning structure towards accurate 3D prompts generation.

%% file: secs/3_method.tex
\begin{figure*}[tb]
  \centering
  \includegraphics[width=0.85\linewidth]{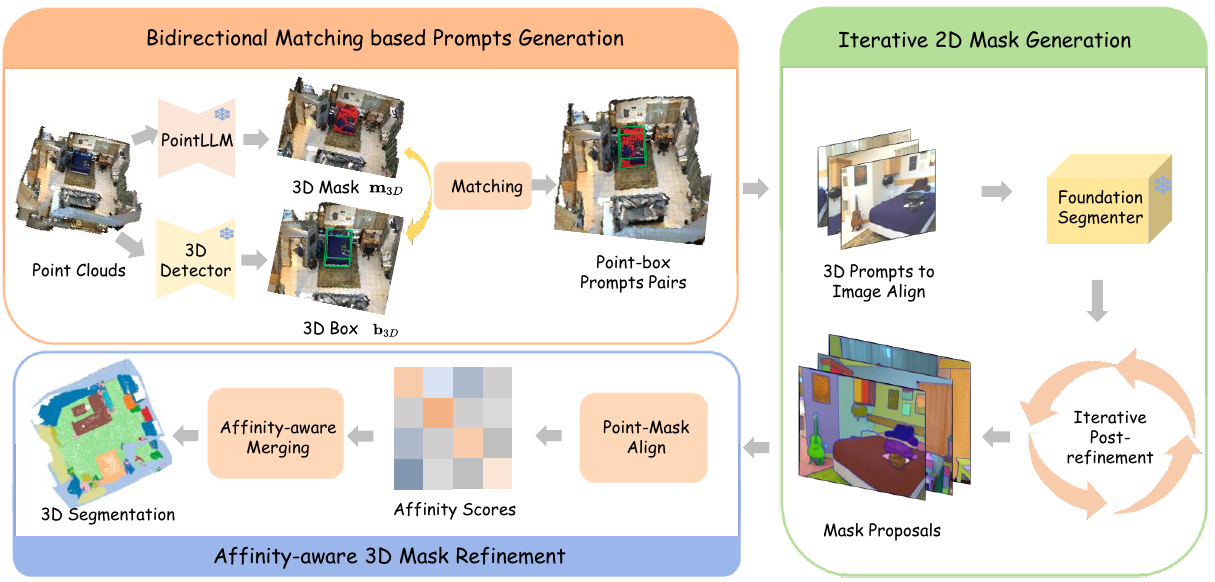}
  \caption{\textbf{Overview of our proposed PointSeg}. Our framework requires no training and aims to segment anything in 3D scene via three distinct stages: Bidirectional Matching based Prompts Generation, Iterative 2D Mask Generation, and Affinity-aware 3D Mask Refinement.
  }
  \label{fig:2}
\end{figure*}
\section{Method}

As illustrated in ~\cref{fig:2}, we build a training-free 3D scene segmentation framework based on the off-the-shelf foundation models. Our PointSeg consists of three parts: 
1) Bidirectional Matching based Prompts Generation (BMP) (Section~\ref{sect:BMP}). Given the reconstructed scene point cloud \begin{math} \mathcal{P} \end{math} $=\{\mathbf{p}\}$ together with a set of posed RGB-D images $\{I_m\}_{m=1}^M$, PointSeg first employs a two-branch prompts learning structure to acquire the 3D mask $\mathbf{m}_{3D}$ and box $\mathbf{b}_{3D}$prompts, respectively. And the final point-box prompts pairs are obtained by the further bidirectional matching. These prompts serve as inputs to 2D vision foundation models, such as SAM 2~\citep{ravi2024sam}, after aligning with the pixels in 2D images. 
2) Iterative 2D Mask Generation (Section~\ref{sect:IPR}). Then, we perform Iterative Post-refinement (IPR) to enable the generation of mask proposals adaptively. 3) Affinity-aware 3D Mask Refinement (Section~\ref{sect:AM}). We calculate the affinity scores between the points generated in the point-box pairs and the mask proposals, followed by the Affinity-aware Merging (AM) (see Algorithm~\ref{alg:2}) to obtain the final 3D segmentation masks.

\subsection{Bidirectional Matching based Prompts Generation}
\label{sect:BMP}
Towards the generation of accurate 3D prompts, we dive into the exploration of  the intrinsic characteristics of 3D data and design a two-branch prompts learning structure. The central concept of our approach entails identifying 3D points and boxes within scenes, serving as inherent 3D prompts, and aligning their projected pixel prompts across various frames. This ensures consistency across frames, both in terms of pixel prompts and the masks predicted by the segmenter.

\noindent\textbf{Two-branch Prompts Generation.} 
Inspired by the strong ability of semantic understanding of some works~\citep{xu2023pointllm,zhang2022pointclip,zhu2023pointclip} in 3D scene, we first employ PointLLM~\citep{xu2023pointllm} for localization and rough segmentation in the upper branch in BMP of Figure~\ref{fig:2}. Given the point cloud \begin{math} \mathcal{P} \end{math} $\in \mathbb{R} ^{N\times3}$, we also apply realistic projection to generate the different $S$ views, using the zero-initialized 3D grid $G\in \mathbb{R} ^{H\times W \times D}$, where $H$/$W$ denote the spatial resolutions and $D$ represents the depth dimension vertical to the view plane. For each view, the normalized 3D coordinates of the input point cloud $\mathbf{p} = (x,y,z)$ in a voxel in the grid can be denoted as 
\begin{equation}
        G(\lceil sHx \rceil,\lceil sWy \rceil,\lceil Dz \rceil )=z,
\end{equation}
where $s\in (0, 1]$ is the scale factor to adjust the projected shape size. Following PointCLIPv2, we further apply the quantize, densify, smooth, and squeeze operations to obtain the projected depth maps $V =\{v_i\}_{i=1}^S$. For the textual input, we utilize the large-scale language models~\citep{brown2020language} to obtain a series of 3D-specific descriptions. After feeding the depth maps and texts into their respective encoders, we can obtain the dense visual features $\{f_i\}_{i=1}^S$ where $f_i\in \mathbb{R} ^{H\times W \times C}$ and the text feature $f_t \in \mathbb{R} ^{K \times C}$. Then, we segment different parts of the shape on multi-view depth maps by dense alignment for each view $i$ and average the back-projected logits of different views into the 3D space, formulated as: 
\begin{equation}
        f_m = average(Proj^{-1}(f_i \cdot f_t^T)), 
\end{equation}
where $f_m$ is the segmentation logits in 3D space.

Apart from the point-level prompts in the 3D mask, we intend to inquire into how to generate dense region-level prompts to fully unleash the advantages of 3D prompts in the another branch. To enhance the ability to segment regions accurately in the subsequent segmenters, we exploit the localization abilities of 3D detector to provide more explicit prompts in the form of bounding boxes. The point cloud \begin{math} \mathcal{P} \end{math} is taken as input by the frozen 3D detectors to generate the 3D bounding box $(x,y,z,w,h,l)$ for each category with corresponding proposal features $f_b$, which can be represented as 
\begin{equation}
        f_b = Det(\mathcal{P}).
\end{equation}

\begin{figure}[tb]
  \centering
  \includegraphics[width=0.95\linewidth]{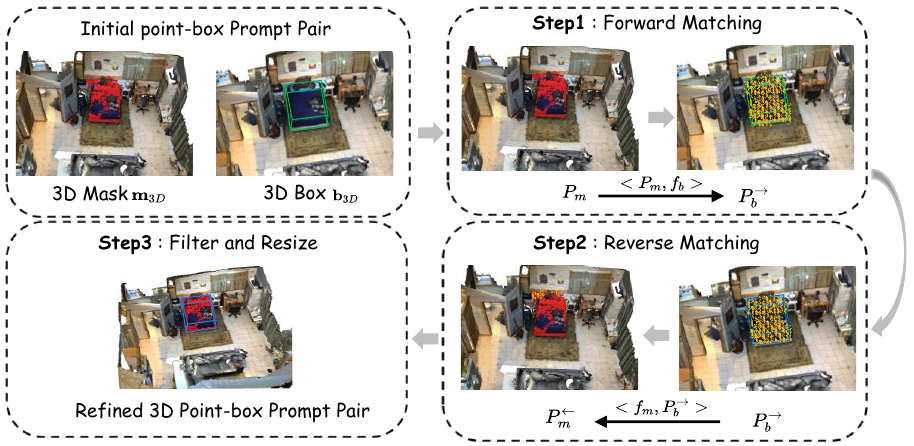}
  \caption{\textbf{Illustration of the proposed bidirectional matching}, which consists of three steps: Forward matching, Reserve matching, and Filter and Resize.
  }
  \label{fig:3}
\end{figure}
\noindent\textbf{Bidirectional Matching.} 
Given the coarse segmentation mask and the bounding box, we can already conduct the alignment to 2D images. However, the naive prompts often result in inaccurate and fragmented outcomes, riddled with numerous outliers. Therefore, we design the bidirectional matching strategy to put constraints on the point and box for high quality promptable segmentation.

Considering the extracted features $f_m$ and $f_b$, we compute the region-wise similarity between the two features to discovery the best matching locations 
\begin{equation}\label{eq:capt}
     <f_m^i,f_b^j> =\frac{f_m^i \cdot f_b^j}{\Vert f_m^i \Vert \cdot \Vert f_b^j \Vert} ,
\end{equation}
where $<f_m^i,f_b^j>$ denotes the cosine similarity between $i$-th mask feature and $j$-th box feature. Ideally, the matched regions should have the highest similarity. Then as illustrated in ~\cref{fig:3}, we propose to eliminate the matching outliers based on the similarity scores in three steps:
\begin{itemize}
    \item First, we compute the forward similarity $<P_m,f_b>$ between the points in the mask $P_m$ and the box $f_b$. Using this score, the bipartite matching is performed to acquire the forward matched points $P_b^{\rightarrow}$ within the box.
    \item Then, we perform reverse matching between the matched points $P_b^{\rightarrow}$ and $f_m$ to obtain the reverse matched points $P_m^{\leftarrow}$ within the mask using the reverse similarity $<f_m,P_b^{\rightarrow}>$.
    \item Finally, we resize the box according to the points in the forward sets if the corresponding reverse points are not within mask, denoted as  $\hat{P}_b=\{ \mathbf{p}_b^i \in P_b^{\rightarrow} | \mathbf{p}_m^i \ in \ 
 \mathbf{m}_{3D}\}$. Similarly,  we adjust points in the mask by filtering out the points in the reverse set if the corresponding forward points are not within the box, denoted as  $\hat{P}_m=\{ \mathbf{p}_m^i \in P_m^{\leftarrow} | \mathbf{p}_b^i \ in \ \mathbf{b}_{3D}\}$.
\end{itemize}
In this way, we can form the point-box pairs with a new set of points in the mask and a different box with new size, which are feed into the further segmentation module.

\begin{figure}[htb]
  \centering
   \begin{algorithm}[H]
            \caption{Iterative Post-refinement}
            \label{alg:1}
            \textbf{Input:} the projected point coordinates $\mathbf{x}$ and box $\mathbf{b}$ in frame $i$, the predefined threshold $\vartheta$\\
            \textbf{Output:} the refined mask $M_i$
        \begin{algorithmic}[1] 
        \STATE $\Delta$ $\leftarrow$ $\infty$
        \STATE $i \leftarrow 1$
        \STATE $M_0 = Dec_M(\mathbf{x},\mathbf{b})$
        \WHILE{$\Delta > \vartheta $}
        \STATE $M_i = Dec_M(\mathbf{x},\mathbf{b}, M_{i-1})$
        \STATE $\Delta \leftarrow \frac{\sum_{j=1}^{N}(M_{i,j} - M_{{i-1},j})}{M_{{i-1},j}}$
        \STATE $i \leftarrow i+1$
        \ENDWHILE
        \STATE \textbf{return} $M_i$
    \end{algorithmic}
    \end{algorithm}
    \vspace{-20pt}
\end{figure}
\subsection{Iterative 2D Mask Generation}
\label{sect:IPR}
Conditioned on the reorganized point-box pairs, we then make the alignment between the 3D prompts pairs and the 2D images. In particular, given a point $\mathbf{p}$ in the prompts pairs with the camera intrinsic matrix $I_i$ and world-to-camera extrinsic matrix $E_i$, the corresponding pixel projection $\mathbf{x}$ can be calculated by
\begin{equation}
        \mathbf{x}=(u,v) = I_i \cdot E_i \cdot \tilde{\mathbf{p}} ,
\end{equation}
where $\tilde{\mathbf{p}}$ is the homogeneous coordinates of $\mathbf{p}$. Similarly, the corresponding projected box across images can be denoted as $\mathbf{b}= (u,v, h, w)$.

The vanilla segmentation model, such as SAM 2~\citep{ravi2024sam}, accepts various inputs such as pixel coordinates, bounding boxes or masks and predict the segmentation area associated with each prompt. Hence, we feed the projected 2D point-box pairs calculated before into the foundation segmenters. 

\begin{figure}[htb]
  \centering
   \begin{algorithm}[H]
            \caption{Affinity-aware Merging}
            \label{alg:2}
            \textbf{Input:} affinity matrix $A \in \mathbb{R}^{N \times N}$ where $A_{i,j}$ indicates the affinity score between two points $p_i$ and $p_j$, and $N$ is the number of points. \\
            \textbf{Output:} mask label $l \in \mathbb{R}^{N}$.
            \begin{algorithmic}[1]
            \STATE $l \leftarrow 0$, $id \leftarrow 1$
            \FOR{$i \leftarrow 1$ to $N$}
            \IF{$l_i = 0$}
            \STATE $l_i \leftarrow id$
            \FOR{each $j$ in neighbors of $i$}
            \IF{$l_j \neq 0$}
            \STATE continue
            \ENDIF
            \STATE $R \leftarrow \{p_k | l_k = id\}$
            \STATE $A_{R,p_j} \leftarrow$ region-point score$(R, p_j , A)$
            \IF{$A_{R,p_j} > \tau$}
            \STATE $l_j \leftarrow id$, $i \leftarrow j$, $id \leftarrow id+1$
            \STATE continue
            \ENDIF
            \ENDFOR
            \STATE $id \leftarrow id + 1$
            \ENDIF
            \ENDFOR
            \end{algorithmic}
        \end{algorithm}
        \vspace{-20pt}
\end{figure}

\begin{figure*}[htb]
  \centering
   \includegraphics[width=0.85\linewidth]{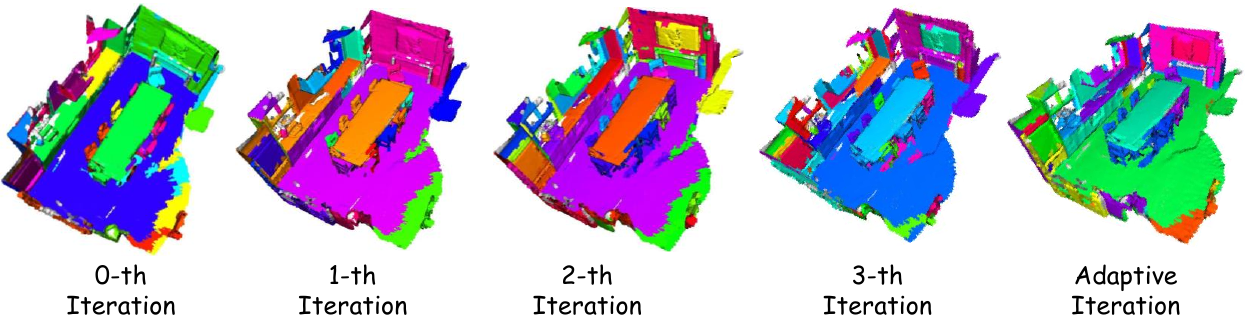}
   \caption{\textbf{Different Iteration strategies.} Without adaptive iteration, the segmentation results can be sensitive to the number of fixed iteration steps. But the performance of adaptive iteration is proved to be more effective. }
    \label{fig4}
\end{figure*}
\noindent\textbf{Iterative Post-refinement.}
Through the above operation, we can obtain 2D segmentation mask on all frames from the decoder, which however, might include rough edges and isolated background noises. For further refinement, we iteratively feed the mask back into the decoder $Dec_M$ for the adaptive post-processing. As illustrated in Algorithm~\ref{alg:1}, we first obtain the 2D mask $M_0$ by feeding the 2D point-box pairs into the SAM-based decoder. Then we prompt the decoder additionally with this mask along with these projected prompt pairs to obtain the next mask. And the initial value $\Delta$ to record change is set to infinity. In each subsequent iteration, we calculate the change ratio between the two adjacent masks and compare it with our predefined threshold $\vartheta$, which is set to 5$\%$ by default. We repeat this iterative process until the change value falls below the threshold adaptively. It is worth noting that we have also tried the method of fixed iteration steps, but this adaptive iteration method have proved to be more effective which is shown in ~\cref{fig4} and Table~\ref{tab:iterative}.

\subsection{Affinity-aware 3D Mask Refinement}
\label{sect:AM}
After previous procedures, we have obtained the final set of 2D segmentation masks across frames. With the ultimate goal of segmenting all points in the 3D scene, we employ the affinity-aware merging algorithm to generate the final 3D masks.

\noindent\textbf{Affinity-aware Merging.}
Based on the $i$-th projected point-box pair on the $m$-th image and their 2D image segmentation mask, we compute the normalized distribution of the mask labels, denoted as $\mathbf{d}_{i,m}$. The affinity score between two projected points in the $m$-th image can be computed by the cosine similarity between the two distributions, which can be represented as:
\begin{equation}\label{eq:capt}
     A_{i,j}^m =\frac{\mathbf{d}_{i,m} \cdot \mathbf{d}_{j,m}}{\Vert \mathbf{d}_{i,m} \Vert \cdot \Vert \mathbf{d}_{j,m} \Vert}.
\end{equation}
The final affinity score across different images can be computed by the weighted-sum:
\begin{equation}\label{eq:7}
     A_{i,j} =\frac{\sum_{m=1}^M \alpha_{i,j}^m \cdot A_{i,j}^m}{\sum_{m=1}^M \alpha_{i,j}^m} ,
\end{equation}
where $\alpha_{i,j}^m \in (0,1)$ denotes whether it is visible in the images.

Further, we utilize the designed affinity-aware merging algorithm to gradually merge 3D masks using the computed affinity matrix. As illustrated in Algorithm~\ref{alg:2}, the algorithm works on an affinity matrix representing the affinity scores between pairs of points. The goal is to assign labels to these points based on their affinities.

We start by initializing labels and an identifier. It then iterates over each point. If a point hasn't been labeled yet, it gets the current identifier. The algorithm then checks the point's neighbors. If a neighbor is unlabeled, it calculates an affinity score between the set of already labeled points and the current neighbor. If this score surpasses a certain threshold, the neighbor is labeled with the current identifier, the identifier is incremented, and the algorithm continues with this neighbor as the current point. The region-point merging is similar to Equation~\ref{eq:7}, which is computed as the weighted average between the current point and the points inside the region. The whole process repeats until all points have been labeled, effectively grouping points based on their mutual affinities. 

%% file: secs/4_experiment.tex
\section{Experiment}
\subsection{Setup}
\label{s4.1}
\begin{table*}[t!]
\tiny
\vspace{-0.2cm}
\centering
\begin{adjustbox}{width=\linewidth}
	\begin{tabular}{l c|c c c| c c c| c c c }
	\toprule[0.5pt]
    \multirow{2}{*}{Method} & \multirow{2}{*}{Type} & \multicolumn{3}{c|}{ScanNet} &\multicolumn{3}{c|}{ScanNet++} &\multicolumn{3}{c}{KITTI-360}\\
 & & mAP & AP$_{50}$ & AP$_{25}$ &mAP & AP$_{50}$ & AP$_{25}$ & mAP & AP$_{50}$ & AP$_{25}$  \\
 \cmidrule(lr){1-2}\cmidrule(lr){3-5} \cmidrule(lr){6-8} 
  \cmidrule(lr){9-11}
  \multicolumn{11}{l}{\emph{With Training}} \\
  SAM-graph~\citep{guo2023sam} & \emph{specialist model} & 15.1 & 33.3& 59.1& 12.9& 25.3& 43.6&14.7 & 28.0& 43.2\\
  Mask3D~\citep{schult2023mask3d}& \emph{specialist model} & 26.9 & 44.4  & 57.5& 8.8 & 15.0 & 22.3 & 0.1 & 0.4 & 4.2\\
  OpenDAS~\citep{yilmaz2024opendas} & \emph{specialist model} & 28.3 & 49.6  & 66.2 & 19.2 & 35.5 & 52.6 & 20.1 & 32.4 & 52.2\\
  PointSAM~\citep{zhou2024point} &\emph{specialist model} & 32.9 & 56.4  & 72.5& 25.8 & 38.0 & 59.3 & 25.1 & 38.4 & 56.2\\
  \cmidrule(lr){1-11}
  \multicolumn{11}{l}{\emph{Training-free}} \\
    SAM3D~\citep{yang2023sam3d} &\emph{specialist model} & 13.7& 29.7& 54.5& 8.3& 17.5& 33.7& 6.3& 16.0& 35.6\\
    SAI3D~\citep{yin2024sai3d} & \emph{specialist model}& 18.8&42.5  & 62.3& 17.1& 31.1& 49.5&16.5 & 30.2 & 48.6\\
    SAMPro3D~\citep{xu2023sampro3d} &\emph{specialist model} & 22.2 &45.6  & 65.7& 18.9 & 33.7& 51.6& 17.3& 31.1 & 49.6\\
        \cmidrule(lr){1-11}
        \rowcolor{pink!12} \bf PointSeg (Ours) &\emph{generalist model} & \bf{38.5} & \bf{63.6}& \bf{82.1} & \bf{33.8} & \bf{49.1} & \bf{67.2}& \bf{32.3}& \bf{47.2}& \bf{66.5}\\
	\bottomrule[0.5pt]
	\end{tabular}
\end{adjustbox}
\vspace{-8pt}
\caption{\textbf{Results of 3D segmentation on ScanNet, ScanNet++, and KITTI-360 datasets}. We report the mAP and AP scores on the three datasets. Best results are highlighted in \textbf{bold}.}
 \label{tab:1}
\end{table*}

\begin{table*}[t!]
\tiny
\vspace{-0.2cm}
\centering
\begin{adjustbox}{width=\linewidth}
	\begin{tabular}{l|c c c| c c c| c c c }
	\toprule[0.5pt]
    \multirow{2}{*}{Method} & \multicolumn{3}{c|}{ScanNet} &\multicolumn{3}{c|}{ScanNet++} &\multicolumn{3}{c}{KITTI-360}\\
 & mAP & AP$_{50}$ & AP$_{25}$ &mAP & AP$_{50}$ & AP$_{25}$ & mAP & AP$_{50}$ & AP$_{25}$  \\
 \cmidrule(lr){1-1}\cmidrule(lr){2-4} \cmidrule(lr){5-7} 
  \cmidrule(lr){8-10}
   \rowcolor{pink!12} + SAM 2~\citep{ravi2024sam} & \bf{38.5} & \bf{63.6}& \bf{82.1} & \bf{33.8} & \bf{49.1} & \bf{67.2}& \bf{32.3}& \bf{47.2}& \bf{66.5}  \\
   + SAM~\citep{kirillov2023segment} & 36.3 & 60.2 & 79.3 & 31.2 & 46.5 & 64.8& 29.9& 44.5 & 63.3  \\
    + MobileSAM~\citep{zhang2023faster} & 26.2 & 49.8 & 68.3 & 19.6 & 36.4 & 55.2 & 20.6 & 34.6 & 53.3  \\
  + FastSAM~\citep{zhao2023fast} & 26.9 & 50.8 & 69.1 & 20.5 & 37.7 & 56.5 & 21.2 & 35.8 & 54.5 \\
  + EfficientSAM~\citep{xiong2023efficientsam} & 33.5 & 57.2 & 75.8 & 27.8 & 43.6 & 62.4 & 27.5 & 41.7 & 60.6  \\
	\bottomrule[0.5pt]
	\end{tabular}
\end{adjustbox}
\caption{\textbf{Results of integrating with different segmentation models} on ScanNet, ScanNet++, and KITTI-360 datasets.}
 \label{tab:2}
 \vspace{-8pt}
\end{table*}

\paragraph{Baselines.}
We compare our approach with both training-based and training-free baselines. For training-based comparison, we select the state-of-the-art transformer-based
method PointSAM~\citep{zhou2024point}. For training-free methods, we compare against the 2D-to-3D lifting methods~\citep{yang2023sam3d,xu2023sampro3d} and the 3D-to-2D projection methods~\citep{xu2023sampro3d} respectively.  For Implementation, we apply the V-DETR~\citep{shen2023v} and VirConv~\citep{wu2023virtual} as the indoor and outdoor 3D detector.
\begin{table}[ht]
\tiny
\resizebox{\linewidth}{!}{
\begin{tabular}{lccc}
	\toprule[0.5pt]
        Method &  mAP & AP$_{50}$ & AP$_{25}$ \\
		\cmidrule(lr){1-4}
             + PointCLIP~\citep{zhang2022pointclip} & 32.3 & 56.9 & 76.2 \\
             + PointCLIPv2~\citep{zhu2023pointclip}  & 34.6 &58.2 &77.6 \\
             + ULIP~\citep{xue2023ulip}  &  34.1 & 57.8 & 77.3 \\
            + ULIPv2~\citep{xue2024ulip}  & 35.7 &59.1 &78.1 \\
            + PointBIND~\citep{guo2023point}  &  36.1 & 59.6 & 78.8 \\
           \rowcolor{pink!12} + PointLLM~\citep{xu2023pointllm}  &  36.3 & 60.2 & 79.3 \\
	  \bottomrule[0.5pt]
	\end{tabular}
}
\vspace{-8pt}
\caption{%
    \textbf{Ablations results of different 3D point models}.
}
\label{tab:t3}
\vspace{-10pt}
\end{table}

\begin{table}[ht]
\tiny
\resizebox{\linewidth}{!}{
\begin{tabular}{lccc}
	\toprule[0.5pt]
        Method &  mAP & AP$_{50}$ & AP$_{25}$ \\
		\cmidrule(lr){1-4}
            \multicolumn{4}{l}{\emph{Indoor}} \\
            \rowcolor{pink!12} + V-DETR~\citep{shen2023v} & 36.3 & 60.2 & 79.3 \\
             + Swin3d~\citep{yang2023swin3d}  & 35.6 &59.7 &78.1 \\
             + CAGroup3D~\citep{wang2022cagroup3d}  &  34.2 & 58.1 & 77.2 \\
            \cmidrule(lr){1-4}
            \multicolumn{4}{l}{\emph{Outdoor}} \\
           \rowcolor{pink!12} + Virconv~\citep{wu2023virtual}  & 36.3 & 60.2 & 79.3 \\
            + TED~\citep{wu2023transformation}  &  33.5 & 56.8 & 75.1 \\
            + LoGoNet~\citep{li2023logonet}  &  33.1 &56.2 &74.8 \\
	  \bottomrule[0.5pt]
	\end{tabular}
}
\vspace{-10pt}
\caption{%
    \textbf{Ablations results of different 3D detectors}.
}
\label{tab:t4}
\vspace{-13pt}
\end{table}

\subsection{Main Results}
\label{s4.2}
\noindent\textbf{Comparisons with the state-of-the-art Methods.}
We compare the segmentation results on ScanNet, ScanNet++, and KITTI-360 datasets, covering both indoor and outdoor scenes. As shown in~\cref{tab:1}, comparing to previous training-free methods, our PointSeg obtains 16.3$\%$ mAP, 18$\%$ AP$_{50}$, and 16.4$\%$ AP$_{25}$ performance gains on ScanNet. On the more challenging indoor dataset ScanNet++, our method still obtains 14.9$\%$ mAP, 15.4$\%$ AP$_{50}$, and 15.6$\%$ AP$_{25}$ improvements. Furthermore, when evaluating the performance of our PointSeg on outdoor KITTI-360, our method still surpasses corresponding zero-shot method by 15$\%$ mAP, 16.1$\%$ AP$_{50}$, and 16.9$\%$ AP$_{25}$, respectively. In this regard, our PointSeg demonstrates superior generalization ability to complex 3D scenarios.

Notably, when compared to previous training-based methods, PointSeg outperforms PointSAM~\citep{zhou2024point} by 5.6$\%$-8$\%$, 7.2$\%$-11.1$\%$, 7.9$\%$-10.3$\%$ in terms of mAP, AP$_{50}$, and AP$_{25}$ across various datasets. This further demonstrates the robustness and effectiveness of our approach in 3D segmentation task.

\noindent\textbf{Qualitative Results.}
The representative quantitative segmentation results of our proposed PointSeg on the three datasets are shown in~\cref{fig::motivation} . We also present the quantitative results of the state-of-the-art training-based methods PointSAM~\citep{zhou2024point} and training-free method SAMPro3D~\citep{xu2023sampro3d}. We can observe that PointSAM and SAMPro3D often mistakenly segment an object into two different objects and exhibit poor performance in segmenting relative objects lacked spatial structure. Our PointSeg can handle complex scenes and is capable of generating clean segments on objects of small size, further underscoring the effectiveness of our approach.

\subsection{Ablation Study}
\label{s4.4}
In this section, we conduct extensive ablation studies on ScanNet to show the effectiveness of each component and design. Unless otherwise specified, SAM~\citep{kirillov2023segment} is used as 2D segmentation foundation model for ablation studies by default.

\noindent\textbf{Different Foundation Models.} 
\emph{i)} Apart the basic segmentation foundation model SAM 2~\citep{ravi2024sam}, we also integrate different segmentation foundation model, \emph{i.e.}, SAM~\citep{kirillov2023segment}, MobileSAM~\citep{zhang2023faster}, FastSAM~\citep{zhao2023fast}, and EfficientSAM~\citep{xiong2023efficientsam}, into our framework. As shown in~\cref{tab:2}, PointSeg demonstrates consistent performance improvements among different datasets, where the original SAM 2-based result performs best. This is consistent with the relative results of these methods in 2D segmentation. \emph{ii)} In~\cref{tab:t3}, we change the model for localization of point prompts. The results demonstrate that more accurate points for 3D prompts can indeed contribute to the performance gain. \emph{iii)} In~\cref{tab:t4}, the improvement of different 3D detectors will also bring about the improvement of the performance of our method. These results suggest that our framework can serve as a foundational structure, capable of integrating a variety of fundamental models. And the enhancements observed in these models can be smoothly translated to 3D space, thereby augmenting the overall performance.

\begin{table}[ht]
\tiny
\resizebox{\linewidth}{!}{
\begin{tabular}{cccccc}
	\toprule[0.5pt]
        BMP & IPR & AM & mAP & AP$_{50}$ & AP$_{25}$ \\
		\cmidrule(lr){1-6} 
        -  & -  & -  & 16.2 &40.6 &60.3 \\
        - & -  & \ding{51}  & 28.6 &50.9 &69.7 \\
        - & \ding{51}  & -  &32.5 &55.2 &74.3  \\
        \ding{51} & -  & -  & 29.5 &51.9 &71.2 \\
      \cmidrule(lr){1-6}
        \ding{51} & \ding{51}  & -  & 33.9 &57.8 &76.5 \\
        \ding{51} & -  & \ding{51}  & 31.7 &53.3 &75.1 \\
        - & \ding{51}  & \ding{51}  &34.6 &58.2 &77.5  \\
         \rowcolor{pink!12} \ding{51} & \ding{51}  & \ding{51}  & 36.3 & 60.2 & 79.3 \\
	  \bottomrule[0.5pt]
	\end{tabular}
}
\vspace{-10pt}
\caption{%
    \textbf{Ablations of main components in our framework}.
}
\label{tab:t5}
\vspace{-10pt}
\end{table}
\begin{figure}[t]
    \centering
    \includegraphics[width=1\linewidth]{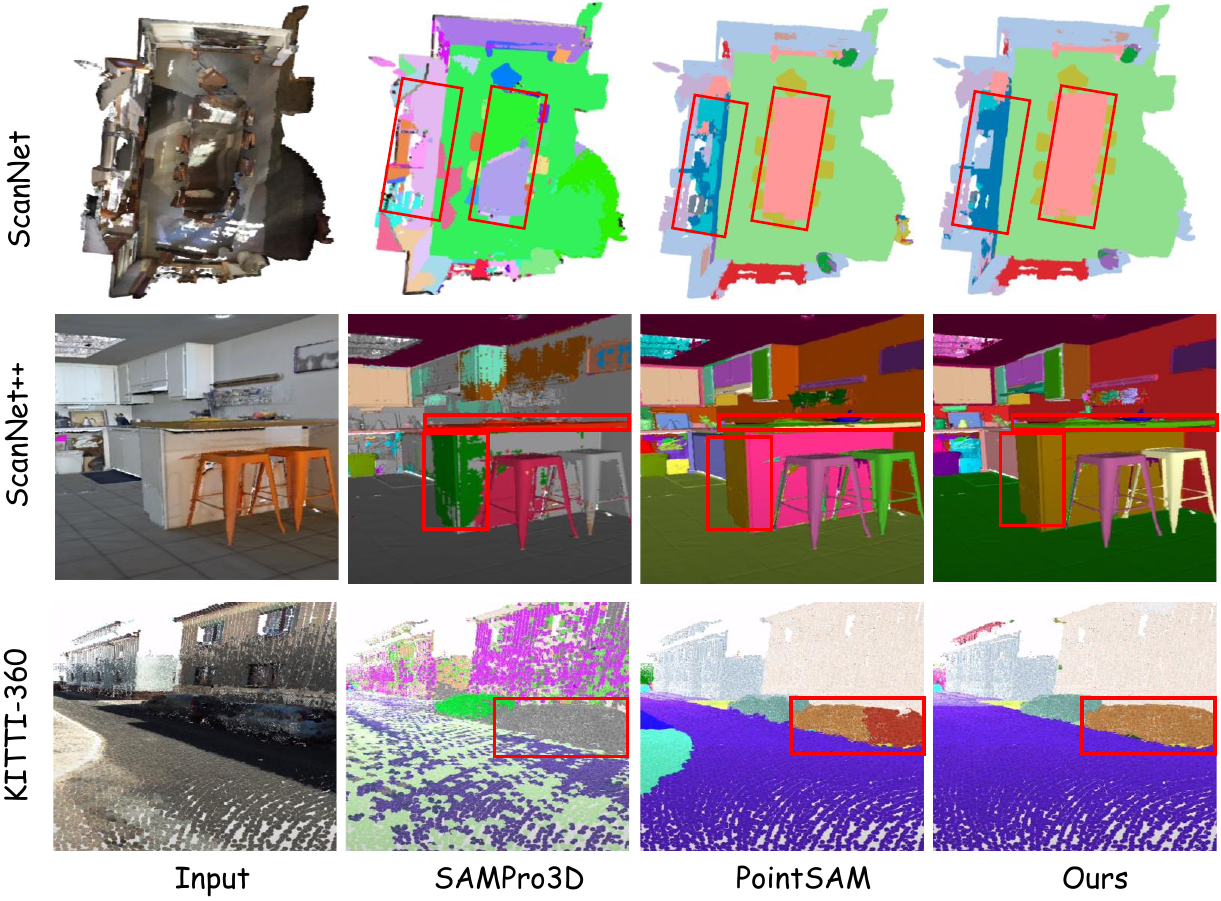}
    \caption{\textbf{More qualitative results comparison} on ScanNet, ScanNet++ and KITTI-360 datasets with comparison to the training-based method PointSAM~\citep{zhou2024point} and the training-free method SAMPro3D~\citep{xu2023sampro3d}.}
    \label{fig::motivation2}
    \vspace{-15pt}
\end{figure}
\noindent\textbf{Main Components.}
Further, we explore the effects of bidirectional matching based prompts generation (BMP), iterative post-refinement (IPR) and affinity-aware merging (AM). We illustrate the importance of different components by removing some parts and keeping all the others unchanged. The baseline setting is to use points and boxes independently as 3D prompts. And the masks from the 2D segmentation model decoder are used to merge the final 3D masks, without any post-refinement. The merging strategy is same as the adjacent frame merging in ~\citep{yang2023sam3d}. The results of components ablations are shown in~\cref{tab:t5}. We observe that using the iterative post-refinement strategy leads to a noticeable increase in performance, which demonstrates the necessity of refining the initial 2D masks. 
The performance degradation caused by the absence of bidirectional matching proves that the constraints between the point and box prompts can indeed help to generate the accurate point-box pairs. And the performance drop without the affinity-aware merging shows that the affinity score is indeed useful to link the point and the masks. 

\begin{table}[ht]
\tiny
\resizebox{\linewidth}{!}{
\begin{tabular}{lcccc}
	\toprule[0.5pt]
        \multicolumn{2}{c}{Strategy} &  mAP & AP$_{50}$ & AP$_{25}$ \\
		\cmidrule(lr){1-5}
      \multicolumn{1}{c}{\multirow{6}{*}{Iter}} & 0 & 30.3 &54.8 & 73.8 \\
            \multicolumn{1}{c}{} &  1 & 31.9 &55.6 & 74.1 \\
            \multicolumn{1}{c}{} &  2  & 32.6 & 56.8 & 75.7 \\
            \multicolumn{1}{c}{} &  3  & 34.7 & 58.5 & 77.8 \\
            \multicolumn{1}{c}{} &  4  & 32.3 & 56.9 & 75.5 \\
            \multicolumn{1}{c}{} &  5  & 30.4 & 54.1 & 73.2 \\
            \cmidrule(lr){1-5}
            \rowcolor{pink!12} \multicolumn{2}{c}{Adaptive} & 36.3 & 60.2 & 79.3 \\
	  \bottomrule[0.5pt]
	\end{tabular}
}
\vspace{-10pt}
\caption{%
    \textbf{Ablation of iterative post-refinement}.
}
\label{tab:iterative}
\vspace{-10pt}
\end{table}

\noindent\textbf{Iterative Post-refinement.}
As mentioned before, when performing the post-refinement during the initial 2D mask generation, we have tried different strategies of fixed numbers method and adaptive iteration method. As shown in~\cref{tab:iterative}, the six rows in the middle represent the method to use a fixed number of iterations and the last row is the adaptive iteration. We can observe that as the number of iterations increases, the corresponding AP value also becomes higher compared to no iteration, which shows that the obtained mask is also more accurate. In the method with a fixed number of iterations, the results reach the highest in the third iteration, but are still lower than those in the adaptive iteration method. This largely illustrates the effectiveness of the iterative post-processing method in generating more accurate masks, and also reveals that the adaptive iteration is more beneficial.


\begin{table}[ht]
\tiny
\resizebox{\linewidth}{!}{
	\begin{tabular}{lccc}
	\toprule[0.5pt]
		Strategy &  mAP & AP$_{50}$ & AP$_{25}$ \\
		\cmidrule(lr){1-4}
      \emph{no}  &  22.2 &49.6 &67.3 \\
            forward &   30.9 & 55.5 & 73.5 \\
            reverse &   32.6 & 56.7 & 75.5  \\
            \rowcolor{pink!12} bidirectional & 36.3 & 60.2 & 79.3 \\
	  \bottomrule[0.5pt]
	\end{tabular}
}
\vspace{-10pt}
\caption{%
    \textbf{Ablation of bidirectional matching}. \emph{no} means no matching.
}
\label{tab:Matching}
\vspace{-10pt}
\end{table}
\noindent\textbf{Different Matching Strategy.}
To validate the effect of different matching method, we explore the effects of the forward matching and the reverse matching of the proposed bidirectional matching, as shown in~\cref{tab:Matching}. Without the guidance from the respective point and box, the naive prompts contain many invalid points and regions, which provide negative prompts for the following segmentation models. Our bidirectional matching improves the performance of forward and reverse matching by 5.4$\%$ and 3.7 $\%$, which show the effectiveness of the proposed bidirectional matching strategy.
\begin{table}[ht]
\tiny
\resizebox{\linewidth}{!}{
	\begin{tabular}{llcc}
	\toprule[0.5pt]
      Strategy &  mAP & AP$_{50}$ & AP$_{25}$ \\
		\cmidrule(lr){1-1} \cmidrule(lr){2-2} \cmidrule(lr){3-3} \cmidrule(lr){4-4} 
      PointClip only & 30.2 &51.2 &72.7  \\
        3D detector only & 30.1 &51.3 &72.6 \\
        combine(w/o matching) &34.6 &58.2 &77.5 \\
        \rowcolor{pink!12} combine(w/ matchfing)  & 36.3 & 60.2 & 79.3 \\
	  \bottomrule[0.5pt]
	\end{tabular}
}
\vspace{-10pt}
\caption{%
    \textbf{Ablation of 3D point pre-trained models from the two branches.}
}
\label{tab:Point Models}
\vspace{-13pt}
\end{table}

\begin{table}[ht]
\tiny
\resizebox{\linewidth}{!}{
\begin{tabular}{cccc}
    \toprule
        module &   bidirectional matching & iterative refinement &  affinity-aware merging  \\
        \midrule
        FPS  & 1.53  & 1.05  &  1.96  \\
        \bottomrule
    \end{tabular}
}
\vspace{-10pt}
\caption{%
    \textbf{Inference Speed}.
}
\label{tab:Inference Speed}
\vspace{-10pt}
\end{table}

\noindent\textbf{3D Point Models.}
As shown in~\cref{tab:Point Models}, the missing of PointLLM/3D detector causes the performance drop and only the matching of combination performs best.


\noindent\textbf{Inference Speed.}
As shown in~\cref{tab:Inference Speed}, we test the running efficiency on NVIDIA V100 GPU and the detailed FPS of each module. The FLOPs of our method is 4.2G.

%% file: secs/5_conclusion.tex
\vspace{-8pt}
\section{Conclusion} 
\label{sec:conclusion}
In this paper, we present PointSeg, a novel training-free framework integrating off-the-shelf vision foundation models for solving 3D scene segmentation tasks. The key idea is to learn accurate 3D point-box prompts pairs to enforce the off-the-shelf foundation models. Combining the three universal components, \emph{i.e.}, bidirectional matching based prompts generation, iterative post-refinement and affinity-aware merging, PointSeg can effectively unleash the ability of various foundation models. Extensive experiments on both indoor and outdoor datasets demonstrate that PointSeg outperforms prior unsupervised methods and even surpass fully-supervised by a large margin, which reveals the  superiority of our model in 3D scene understanding task.

%% file: secs/X_suppl.tex
\clearpage
\renewcommand{\thefigure}{A\arabic{figure}}
\setcounter{figure}{0}
\renewcommand{\thetable}{A\arabic{table}}
\setcounter{table}{0}
\renewcommand{\thesection}{A\arabic{section}}
\setcounter{section}{0}
\maketitlesupplementary

The supplementary material presents the following sections to strengthen the main manuscript:
\begin{itemize}
\item Datasets and Metrics.
\item More Ablation Studies.
\end{itemize}

\section{Datasets and Metrics}
To validate the effectiveness of our proposed PointSeg, we conduct extensive experiments on three popular public benchmarks: ScanNet~\citep{dai2017scannet}, ScanNet++~\citep{yeshwanth2023scannet++}, and KITTI-360~\citep{liao2022kitti}. ScanNet provides RGBD images and 3D meshes of 1613 indoor scenes. ScanNet++ is a recently released indoor dataset with more detailed segmentation masks, serving as a more challenging benchmark for 3D scenarios. It contains 280 indoor scenes with high-fidelity geometry and high-resolution RGB images. KITTI-360 is a substantial outdoor dataset that includes 300 suburban scenes, which comprises 320k images and 100k laser scans. We evaluate ours segmentation performance with the widely-used Average Precision (AP) score. Following ~\citep{schult2023mask3d,dai2017scannet}, we report AP with thresholds of 50$\%$ and 25$\%$ (denoted as AP$_{50}$ and AP$_{25}$) as well as AP averaged with IoU thresholds from 50$\%$ to 95$\%$ with a step size of 5$\%$ (mAP).

\section{More Ablation Studies}
\begin{table}[ht]
\tiny
\resizebox{\linewidth}{!}{
	\begin{tabular}{lccc}
	\toprule[0.4pt]
		 Merging   & mAP & AP$_{50}$ & AP$_{25}$ \\
      \cmidrule(lr){1-4}
      BM  & 27.6 &53.5 &73.7 \\
      PM  & 31.5 &55.7 &76.2 \\
      IDM & 32.1 &56.2 &76.8  \\
     \rowcolor{pink!12} AM  & 36.3 & 60.2 & 79.3  \\
	  \bottomrule[0.4pt]
	\end{tabular}
}
\vspace{-10pt}
\caption{%
    \textbf{Ablation of affinity-aware merging algorithm}.
}
\label{tab:affinity-aware}
\end{table}
\noindent\textbf{Affinity-aware Merging.}
In the final mask merging stage, we have also tried other merging algorithm. As shown in~\cref{tab:affinity-aware}, we compare our affinity-aware merging (AM) with (a) bidirectional merging (BM) from~\citep{yang2023sam3d}, (b) pure merging (PM) without affinity scores, which is simplified from our approach and (c) prompt ID based merging (IDM) from~\citep{xu2023sampro3d}. With other inferior merging method, the performance drops dramatically which shows the superiority of our proposed affinity-aware merging algorithm in solving the mask merging problems in 3D scene.

\begin{table}[ht]
\tiny
\resizebox{\linewidth}{!}{
\begin{tabular}{cccc}
        \toprule[0.4pt]
            Ratio &  mAP & AP$_{50}$ & AP$_{25}$ \\
            \midrule[0.4pt]
              1$\%$   & 33.6 & 57.5 & 76.6 \\
              3$\%$   & 35.2 & 59.3 & 78.7 \\
              \rowcolor{pink!12} 5$\%$   & 36.3 & 60.2 & 79.3 \\
              8$\%$   & 35.8 & 59.7 & 78.1 \\
              10$\%$  & 34.6 & 58.3 & 77.9 \\
              15$\%$  & 32.2 & 56.6 & 75.5 \\
            \bottomrule[0.4pt]
        \end{tabular}
}
\vspace{-10pt}
\caption{%
    \textbf{Ablation of mask change ratio}.
}
\label{tab:ratio}
\end{table}
\noindent\textbf{Mask Change Ratio.}
In the iterative post-refinement module, we set a mask change ratio threshold $\vartheta$ as a condition for stopping the iteration. Here, We show the effect of different ratios on the results. As shown in~\cref{tab:ratio}, the overall results perform best when mask change ratio is set as 5$\%$.

%% file: main.bbl
\begin{thebibliography}{72}
\providecommand{\natexlab}[1]{#1}
\providecommand{\url}[1]{\texttt{#1}}
\expandafter\ifx\csname urlstyle\endcsname\relax
  \providecommand{\doi}[1]{doi: #1}\else
  \providecommand{\doi}{doi: \begingroup \urlstyle{rm}\Url}\fi

\bibitem[Brown et~al.(2020)Brown, Mann, Ryder, Subbiah, Kaplan, Dhariwal, Neelakantan, Shyam, Sastry, Askell, et~al.]{brown2020language}
Tom Brown, Benjamin Mann, Nick Ryder, Melanie Subbiah, Jared~D Kaplan, Prafulla Dhariwal, Arvind Neelakantan, Pranav Shyam, Girish Sastry, Amanda Askell, et~al.
\newblock Language models are few-shot learners.
\newblock \emph{Advances in Neural Information Processing Systems}, 33:\penalty0 1877--1901, 2020.

\bibitem[Chang et~al.(2015)Chang, Funkhouser, Guibas, Hanrahan, Huang, Li, Savarese, Savva, Song, Su, et~al.]{chang2015shapenet}
Angel~X Chang, Thomas Funkhouser, Leonidas Guibas, Pat Hanrahan, Qixing Huang, Zimo Li, Silvio Savarese, Manolis Savva, Shuran Song, Hao Su, et~al.
\newblock Shapenet: An information-rich 3d model repository.
\newblock \emph{arXiv preprint arXiv:1512.03012}, 2015.

\bibitem[Choy et~al.(2019)Choy, Gwak, and Savarese]{choy20194d}
Christopher Choy, JunYoung Gwak, and Silvio Savarese.
\newblock 4d spatio-temporal convnets: Minkowski convolutional neural networks.
\newblock In \emph{Proceedings of the IEEE/CVF conference on computer vision and pattern recognition}, pages 3075--3084, 2019.

\bibitem[Dai et~al.(2017)Dai, Chang, Savva, Halber, Funkhouser, and Nie{\ss}ner]{dai2017scannet}
Angela Dai, Angel~X Chang, Manolis Savva, Maciej Halber, Thomas Funkhouser, and Matthias Nie{\ss}ner.
\newblock Scannet: Richly-annotated 3d reconstructions of indoor scenes.
\newblock In \emph{Proceedings of the IEEE/CVF Conference on Computer Vision and Pattern Recognition}, pages 5828--5839, 2017.

\bibitem[Ding et~al.(2023)Ding, Yang, Xue, Zhang, Bai, and Qi]{ding2023pla}
Runyu Ding, Jihan Yang, Chuhui Xue, Wenqing Zhang, Song Bai, and Xiaojuan Qi.
\newblock Pla: Language-driven open-vocabulary 3d scene understanding.
\newblock In \emph{Proceedings of the IEEE/CVF Conference on Computer Vision and Pattern Recognition}, pages 7010--7019, 2023.

\bibitem[Engelmann et~al.(2020)Engelmann, Bokeloh, Fathi, Leibe, and Nie{\ss}ner]{engelmann20203d}
Francis Engelmann, Martin Bokeloh, Alireza Fathi, Bastian Leibe, and Matthias Nie{\ss}ner.
\newblock 3d-mpa: Multi-proposal aggregation for 3d semantic instance segmentation.
\newblock In \emph{Proceedings of the IEEE/CVF Conference on Computer Vision and Pattern Recognition}, pages 9031--9040, 2020.

\bibitem[Fan et~al.(2021)Fan, Dong, Zhu, Lv, Ye, and Wang]{fan2021scf}
Siqi Fan, Qiulei Dong, Fenghua Zhu, Yisheng Lv, Peijun Ye, and Fei-Yue Wang.
\newblock Scf-net: Learning spatial contextual features for large-scale point cloud segmentation.
\newblock In \emph{Proceedings of the IEEE/CVF Conference on Computer Vision and Pattern Recognition}, pages 14504--14513, 2021.

\bibitem[Ghiasi et~al.(2022)Ghiasi, Gu, Cui, and Lin]{ghiasi2022scaling}
Golnaz Ghiasi, Xiuye Gu, Yin Cui, and Tsung-Yi Lin.
\newblock Scaling open-vocabulary image segmentation with image-level labels.
\newblock In \emph{European Conference on Computer Vision}, pages 540--557. Springer, 2022.

\bibitem[Goyal et~al.(2021)Goyal, Law, Liu, Newell, and Deng]{goyal2021revisiting}
Ankit Goyal, Hei Law, Bowei Liu, Alejandro Newell, and Jia Deng.
\newblock Revisiting point cloud shape classification with a simple and effective baseline.
\newblock In \emph{International Conference on Machine Learning}, pages 3809--3820. PMLR, 2021.

\bibitem[Graham et~al.(2018)Graham, Engelcke, and Van Der~Maaten]{graham20183d}
Benjamin Graham, Martin Engelcke, and Laurens Van Der~Maaten.
\newblock 3d semantic segmentation with submanifold sparse convolutional networks.
\newblock In \emph{Proceedings of the IEEE Conference on Computer Vision and Pattern Recognition}, pages 9224--9232, 2018.

\bibitem[Guo et~al.(2023{\natexlab{a}})Guo, Zhu, Peng, Wang, Shen, Hu, and Zhou]{guo2023sam}
Haoyu Guo, He Zhu, Sida Peng, Yuang Wang, Yujun Shen, Ruizhen Hu, and Xiaowei Zhou.
\newblock Sam-guided graph cut for 3d instance segmentation.
\newblock \emph{arXiv preprint arXiv:2312.08372}, 2023{\natexlab{a}}.

\bibitem[Guo et~al.(2023{\natexlab{b}})Guo, Zhang, Zhu, Tang, Ma, Han, Chen, Gao, Li, Li, et~al.]{guo2023point}
Ziyu Guo, Renrui Zhang, Xiangyang Zhu, Yiwen Tang, Xianzheng Ma, Jiaming Han, Kexin Chen, Peng Gao, Xianzhi Li, Hongsheng Li, et~al.
\newblock Point-bind \& point-llm: Aligning point cloud with multi-modality for 3d understanding, generation, and instruction following.
\newblock \emph{arXiv preprint arXiv:2309.00615}, 2023{\natexlab{b}}.

\bibitem[Han et~al.(2020)Han, Zheng, Xu, and Fang]{han2020occuseg}
Lei Han, Tian Zheng, Lan Xu, and Lu Fang.
\newblock Occuseg: Occupancy-aware 3d instance segmentation.
\newblock In \emph{Proceedings of the IEEE/CVF Conference on Computer Vision and Pattern Recognition}, pages 2940--2949, 2020.

\bibitem[He et~al.(2022)He, Chen, Xie, Li, Doll{\'a}r, and Girshick]{he2022masked}
Kaiming He, Xinlei Chen, Saining Xie, Yanghao Li, Piotr Doll{\'a}r, and Ross Girshick.
\newblock Masked autoencoders are scalable vision learners.
\newblock In \emph{Proceedings of the IEEE/CVF conference on computer vision and pattern recognition}, pages 16000--16009, 2022.

\bibitem[He et~al.(2024)He, Peng, Jiang, Wu, Ji, Zhang, Wang, Wang, Chen, and Wu]{he2024unim}
Qingdong He, Jinlong Peng, Zhengkai Jiang, Kai Wu, Xiaozhong Ji, Jiangning Zhang, Yabiao Wang, Chengjie Wang, Mingang Chen, and Yunsheng Wu.
\newblock Unim-ov3d: Uni-modality open-vocabulary 3d scene understanding with fine-grained feature representation.
\newblock In \emph{International Joint Conference on Artificial Intelligence}, 2024.

\bibitem[Hou et~al.(2019)Hou, Dai, and Nie{\ss}ner]{hou20193d}
Ji Hou, Angela Dai, and Matthias Nie{\ss}ner.
\newblock 3d-sis: 3d semantic instance segmentation of rgb-d scans.
\newblock In \emph{Proceedings of the IEEE/CVF Conference on Computer Vision and Pattern Recognition}, pages 4421--4430, 2019.

\bibitem[Hu et~al.(2020)Hu, Yang, Xie, Rosa, Guo, Wang, Trigoni, and Markham]{hu2020randla}
Qingyong Hu, Bo Yang, Linhai Xie, Stefano Rosa, Yulan Guo, Zhihua Wang, Niki Trigoni, and Andrew Markham.
\newblock Randla-net: Efficient semantic segmentation of large-scale point clouds.
\newblock In \emph{Proceedings of the IEEE/CVF Conference on Computer Vision and Pattern Recognition}, pages 11108--11117, 2020.

\bibitem[Huang et~al.(2023)Huang, Wu, Chen, Zhao, Zhu, and Lasenby]{huang2023openins3d}
Zhening Huang, Xiaoyang Wu, Xi Chen, Hengshuang Zhao, Lei Zhu, and Joan Lasenby.
\newblock Openins3d: Snap and lookup for 3d open-vocabulary instance segmentation.
\newblock \emph{arXiv preprint arXiv:2309.00616}, 2023.

\bibitem[Hui et~al.(2022)Hui, Tang, Shen, Xie, and Yang]{hui2022learning}
Le Hui, Linghua Tang, Yaqi Shen, Jin Xie, and Jian Yang.
\newblock Learning superpoint graph cut for 3d instance segmentation.
\newblock \emph{Advances in Neural Information Processing Systems}, 35:\penalty0 36804--36817, 2022.

\bibitem[Jia et~al.(2021)Jia, Yang, Xia, Chen, Parekh, Pham, Le, Sung, Li, and Duerig]{jia2021scaling}
Chao Jia, Yinfei Yang, Ye Xia, Yi-Ting Chen, Zarana Parekh, Hieu Pham, Quoc Le, Yun-Hsuan Sung, Zhen Li, and Tom Duerig.
\newblock Scaling up visual and vision-language representation learning with noisy text supervision.
\newblock In \emph{International Conference on Machine Learning}, pages 4904--4916. PMLR, 2021.

\bibitem[Jiang et~al.(2020)Jiang, Zhao, Shi, Liu, Fu, and Jia]{jiang2020pointgroup}
Li Jiang, Hengshuang Zhao, Shaoshuai Shi, Shu Liu, Chi-Wing Fu, and Jiaya Jia.
\newblock Pointgroup: Dual-set point grouping for 3d instance segmentation.
\newblock In \emph{Proceedings of the IEEE/CVF Conference on Computer Vision and Pattern Recognition}, pages 4867--4876, 2020.

\bibitem[Jiang et~al.(2022)Jiang, Li, Yang, Gao, Wang, Tai, and Wang]{jiang2022prototypical}
Zhengkai Jiang, Yuxi Li, Ceyuan Yang, Peng Gao, Yabiao Wang, Ying Tai, and Chengjie Wang.
\newblock Prototypical contrast adaptation for domain adaptive semantic segmentation.
\newblock In \emph{European Conference on Computer Vision}, pages 36--54. Springer, 2022.

\bibitem[Kerr et~al.(2023)Kerr, Kim, Goldberg, Kanazawa, and Tancik]{kerr2023lerf}
Justin Kerr, Chung~Min Kim, Ken Goldberg, Angjoo Kanazawa, and Matthew Tancik.
\newblock Lerf: Language embedded radiance fields.
\newblock In \emph{Proceedings of the IEEE/CVF International Conference on Computer Vision}, pages 19729--19739, 2023.

\bibitem[Kirillov et~al.(2023)Kirillov, Mintun, Ravi, Mao, Rolland, Gustafson, Xiao, Whitehead, Berg, Lo, et~al.]{kirillov2023segment}
Alexander Kirillov, Eric Mintun, Nikhila Ravi, Hanzi Mao, Chloe Rolland, Laura Gustafson, Tete Xiao, Spencer Whitehead, Alexander~C Berg, Wan-Yen Lo, et~al.
\newblock Segment anything.
\newblock \emph{arXiv preprint arXiv:2304.02643}, 2023.

\bibitem[Kobayashi et~al.(2022)Kobayashi, Matsumoto, and Sitzmann]{kobayashi2022decomposing}
Sosuke Kobayashi, Eiichi Matsumoto, and Vincent Sitzmann.
\newblock Decomposing nerf for editing via feature field distillation.
\newblock \emph{Advances in Neural Information Processing Systems}, 35:\penalty0 23311--23330, 2022.

\bibitem[Kolodiazhnyi et~al.(2024)Kolodiazhnyi, Vorontsova, Konushin, and Rukhovich]{kolodiazhnyi2024top}
Maksim Kolodiazhnyi, Anna Vorontsova, Anton Konushin, and Danila Rukhovich.
\newblock Top-down beats bottom-up in 3d instance segmentation.
\newblock In \emph{Proceedings of the IEEE/CVF Winter Conference on Applications of Computer Vision}, pages 3566--3574, 2024.

\bibitem[Kundu et~al.(2020)Kundu, Yin, Fathi, Ross, Brewington, Funkhouser, and Pantofaru]{kundu2020virtual}
Abhijit Kundu, Xiaoqi Yin, Alireza Fathi, David Ross, Brian Brewington, Thomas Funkhouser, and Caroline Pantofaru.
\newblock Virtual multi-view fusion for 3d semantic segmentation.
\newblock In \emph{European Conference on Computer Vision}, pages 518--535. Springer, 2020.

\bibitem[Lahoud et~al.(2019)Lahoud, Ghanem, Pollefeys, and Oswald]{lahoud20193d}
Jean Lahoud, Bernard Ghanem, Marc Pollefeys, and Martin~R Oswald.
\newblock 3d instance segmentation via multi-task metric learning.
\newblock In \emph{Proceedings of the IEEE/CVF International Conference on Computer Vision}, pages 9256--9266, 2019.

\bibitem[Li et~al.(2023)Li, Ma, Hou, Shi, Yang, Liu, Wu, Chen, Li, Qiao, et~al.]{li2023logonet}
Xin Li, Tao Ma, Yuenan Hou, Botian Shi, Yuchen Yang, Youquan Liu, Xingjiao Wu, Qin Chen, Yikang Li, Yu Qiao, et~al.
\newblock Logonet: Towards accurate 3d object detection with local-to-global cross-modal fusion.
\newblock In \emph{Proceedings of the IEEE/CVF Conference on Computer Vision and Pattern Recognition}, pages 17524--17534, 2023.

\bibitem[Li et~al.(2018)Li, Bu, Sun, Wu, Di, and Chen]{li2018pointcnn}
Yangyan Li, Rui Bu, Mingchao Sun, Wei Wu, Xinhan Di, and Baoquan Chen.
\newblock Pointcnn: Convolution on x-transformed points.
\newblock \emph{Advances in Neural Information Processing Systems}, 31, 2018.

\bibitem[Liang et~al.(2023)Liang, Wu, Dai, Li, Zhao, Zhang, Zhang, Vajda, and Marculescu]{liang2023open}
Feng Liang, Bichen Wu, Xiaoliang Dai, Kunpeng Li, Yinan Zhao, Hang Zhang, Peizhao Zhang, Peter Vajda, and Diana Marculescu.
\newblock Open-vocabulary semantic segmentation with mask-adapted clip.
\newblock In \emph{Proceedings of the IEEE/CVF Conference on Computer Vision and Pattern Recognition}, pages 7061--7070, 2023.

\bibitem[Liang et~al.(2021)Liang, Li, Xu, Tan, and Jia]{liang2021instance}
Zhihao Liang, Zhihao Li, Songcen Xu, Mingkui Tan, and Kui Jia.
\newblock Instance segmentation in 3d scenes using semantic superpoint tree networks.
\newblock In \emph{Proceedings of the IEEE/CVF International Conference on Computer Vision}, pages 2783--2792, 2021.

\bibitem[Liao et~al.(2022)Liao, Xie, and Geiger]{liao2022kitti}
Yiyi Liao, Jun Xie, and Andreas Geiger.
\newblock Kitti-360: A novel dataset and benchmarks for urban scene understanding in 2d and 3d.
\newblock \emph{IEEE Transactions on Pattern Analysis and Machine Intelligence}, 45\penalty0 (3):\penalty0 3292--3310, 2022.

\bibitem[Liu et~al.(2023)Liu, Zhu, Li, Chen, Wang, and Shen]{liu2023matcher}
Yang Liu, Muzhi Zhu, Hengtao Li, Hao Chen, Xinlong Wang, and Chunhua Shen.
\newblock Matcher: Segment anything with one shot using all-purpose feature matching.
\newblock \emph{arXiv preprint arXiv:2305.13310}, 2023.

\bibitem[Mildenhall et~al.(2021)Mildenhall, Srinivasan, Tancik, Barron, Ramamoorthi, and Ng]{mildenhall2021nerf}
Ben Mildenhall, Pratul~P Srinivasan, Matthew Tancik, Jonathan~T Barron, Ravi Ramamoorthi, and Ren Ng.
\newblock Nerf: Representing scenes as neural radiance fields for view synthesis.
\newblock \emph{Communications of the ACM}, 65\penalty0 (1):\penalty0 99--106, 2021.

\bibitem[Oquab et~al.(2023)Oquab, Darcet, Moutakanni, Vo, Szafraniec, Khalidov, Fernandez, Haziza, Massa, El-Nouby, et~al.]{oquab2023dinov2}
Maxime Oquab, Timoth{\'e}e Darcet, Th{\'e}o Moutakanni, Huy Vo, Marc Szafraniec, Vasil Khalidov, Pierre Fernandez, Daniel Haziza, Francisco Massa, Alaaeldin El-Nouby, et~al.
\newblock Dinov2: Learning robust visual features without supervision.
\newblock \emph{arXiv preprint arXiv:2304.07193}, 2023.

\bibitem[Peng et~al.(2023)Peng, Genova, Jiang, Tagliasacchi, Pollefeys, Funkhouser, et~al.]{peng2023openscene}
Songyou Peng, Kyle Genova, Chiyu Jiang, Andrea Tagliasacchi, Marc Pollefeys, Thomas Funkhouser, et~al.
\newblock Openscene: 3d scene understanding with open vocabularies.
\newblock In \emph{Proceedings of the IEEE/CVF Conference on Computer Vision and Pattern Recognition}, pages 815--824, 2023.

\bibitem[Qi et~al.(2017)Qi, Yi, Su, and Guibas]{qi2017pointnet++}
Charles~Ruizhongtai Qi, Li Yi, Hao Su, and Leonidas~J Guibas.
\newblock Pointnet++: Deep hierarchical feature learning on point sets in a metric space.
\newblock \emph{Advances in Neural Information Processing Systems}, 30, 2017.

\bibitem[Radford et~al.(2021)Radford, Kim, Hallacy, Ramesh, Goh, Agarwal, Sastry, Askell, Mishkin, Clark, et~al.]{radford2021learning}
Alec Radford, Jong~Wook Kim, Chris Hallacy, Aditya Ramesh, Gabriel Goh, Sandhini Agarwal, Girish Sastry, Amanda Askell, Pamela Mishkin, Jack Clark, et~al.
\newblock Learning transferable visual models from natural language supervision.
\newblock In \emph{International Conference on Machine Learning}, pages 8748--8763. PMLR, 2021.

\bibitem[Ravi et~al.(2024)Ravi, Gabeur, Hu, Hu, Ryali, Ma, Khedr, R{\"a}dle, Rolland, Gustafson, et~al.]{ravi2024sam}
Nikhila Ravi, Valentin Gabeur, Yuan-Ting Hu, Ronghang Hu, Chaitanya Ryali, Tengyu Ma, Haitham Khedr, Roman R{\"a}dle, Chloe Rolland, Laura Gustafson, et~al.
\newblock Sam 2: Segment anything in images and videos.
\newblock \emph{arXiv preprint arXiv:2408.00714}, 2024.

\bibitem[Rozenberszki et~al.(2022)Rozenberszki, Litany, and Dai]{rozenberszki2022language}
David Rozenberszki, Or Litany, and Angela Dai.
\newblock Language-grounded indoor 3d semantic segmentation in the wild.
\newblock In \emph{European Conference on Computer Vision}, pages 125--141. Springer, 2022.

\bibitem[Schult et~al.(2023)Schult, Engelmann, Hermans, Litany, Tang, and Leibe]{schult2023mask3d}
Jonas Schult, Francis Engelmann, Alexander Hermans, Or Litany, Siyu Tang, and Bastian Leibe.
\newblock Mask3d: Mask transformer for 3d semantic instance segmentation.
\newblock In \emph{2023 IEEE International Conference on Robotics and Automation (ICRA)}, pages 8216--8223. IEEE, 2023.

\bibitem[Shen et~al.(2024)Shen, Geng, Yuan, Lin, Liu, Wang, Hu, Zheng, and Guo]{shen2023v}
Yichao Shen, Zigang Geng, Yuhui Yuan, Yutong Lin, Ze Liu, Chunyu Wang, Han Hu, Nanning Zheng, and Baining Guo.
\newblock V-detr: Detr with vertex relative position encoding for 3d object detection.
\newblock In \emph{The Twelfth International Conference on Learning Representations}, 2024.

\bibitem[Sun et~al.(2023)Sun, Qing, Tan, and Xu]{sun2023superpoint}
Jiahao Sun, Chunmei Qing, Junpeng Tan, and Xiangmin Xu.
\newblock Superpoint transformer for 3d scene instance segmentation.
\newblock In \emph{Proceedings of the AAAI Conference on Artificial Intelligence}, pages 2393--2401, 2023.

\bibitem[Takmaz et~al.(2023)Takmaz, Fedele, Sumner, Pollefeys, Tombari, and Engelmann]{takmaz2023openmask3d}
Ay{\c{c}}a Takmaz, Elisabetta Fedele, Robert~W Sumner, Marc Pollefeys, Federico Tombari, and Francis Engelmann.
\newblock Openmask3d: Open-vocabulary 3d instance segmentation.
\newblock \emph{arXiv preprint arXiv:2306.13631}, 2023.

\bibitem[Vu et~al.(2022)Vu, Kim, Luu, Nguyen, Kim, and Yoo]{vu2022softgroup++}
Thang Vu, Kookhoi Kim, Tung~M Luu, Thanh Nguyen, Junyeong Kim, and Chang~D Yoo.
\newblock Softgroup++: Scalable 3d instance segmentation with octree pyramid grouping.
\newblock \emph{arXiv preprint arXiv:2209.08263}, 2022.

\bibitem[Wang et~al.(2022)Wang, Ding, Dong, Shi, Li, Li, Li, and Wang]{wang2022cagroup3d}
Haiyang Wang, Lihe Ding, Shaocong Dong, Shaoshuai Shi, Aoxue Li, Jianan Li, Zhenguo Li, and Liwei Wang.
\newblock Cagroup3d: Class-aware grouping for 3d object detection on point clouds.
\newblock \emph{Advances in Neural Information Processing Systems}, 35:\penalty0 29975--29988, 2022.

\bibitem[Wang et~al.(2015)Wang, Li, and An]{wang2015efficient}
Tianyi Wang, Jian Li, and Xiangjing An.
\newblock An efficient scene semantic labeling approach for 3d point cloud.
\newblock In \emph{2015 IEEE 18th International Conference on Intelligent Transportation Systems}, pages 2115--2120. IEEE, 2015.

\bibitem[Wang et~al.(2019)Wang, Sun, Liu, Sarma, Bronstein, and Solomon]{wang2019dynamic}
Yue Wang, Yongbin Sun, Ziwei Liu, Sanjay~E Sarma, Michael~M Bronstein, and Justin~M Solomon.
\newblock Dynamic graph cnn for learning on point clouds.
\newblock \emph{ACM Transactions on Graphics (tog)}, 38\penalty0 (5):\penalty0 1--12, 2019.

\bibitem[Wu et~al.(2023{\natexlab{a}})Wu, Wen, Li, Li, Yang, and Wang]{wu2023transformation}
Hai Wu, Chenglu Wen, Wei Li, Xin Li, Ruigang Yang, and Cheng Wang.
\newblock Transformation-equivariant 3d object detection for autonomous driving.
\newblock In \emph{Proceedings of the AAAI Conference on Artificial Intelligence}, pages 2795--2802, 2023{\natexlab{a}}.

\bibitem[Wu et~al.(2023{\natexlab{b}})Wu, Wen, Shi, Li, and Wang]{wu2023virtual}
Hai Wu, Chenglu Wen, Shaoshuai Shi, Xin Li, and Cheng Wang.
\newblock Virtual sparse convolution for multimodal 3d object detection.
\newblock In \emph{Proceedings of the IEEE/CVF Conference on Computer Vision and Pattern Recognition}, pages 21653--21662, 2023{\natexlab{b}}.

\bibitem[Xiong et~al.(2023)Xiong, Varadarajan, Wu, Xiang, Xiao, Zhu, Dai, Wang, Sun, Iandola, et~al.]{xiong2023efficientsam}
Yunyang Xiong, Bala Varadarajan, Lemeng Wu, Xiaoyu Xiang, Fanyi Xiao, Chenchen Zhu, Xiaoliang Dai, Dilin Wang, Fei Sun, Forrest Iandola, et~al.
\newblock Efficientsam: Leveraged masked image pretraining for efficient segment anything.
\newblock \emph{arXiv preprint arXiv:2312.00863}, 2023.

\bibitem[Xu et~al.(2023)Xu, Yin, Qiu, Liu, Tong, and Han]{xu2023sampro3d}
Mutian Xu, Xingyilang Yin, Lingteng Qiu, Yang Liu, Xin Tong, and Xiaoguang Han.
\newblock Sampro3d: Locating sam prompts in 3d for zero-shot scene segmentation.
\newblock \emph{arXiv preprint arXiv:2311.17707}, 2023.

\bibitem[Xu et~al.(2024)Xu, Wang, Wang, Chen, Pang, and Lin]{xu2023pointllm}
Runsen Xu, Xiaolong Wang, Tai Wang, Yilun Chen, Jiangmiao Pang, and Dahua Lin.
\newblock Pointllm: Empowering large language models to understand point clouds.
\newblock In \emph{Proceedings of the European conference on computer vision (ECCV)}, 2024.

\bibitem[Xu et~al.(2018)Xu, Fan, Xu, Zeng, and Qiao]{xu2018spidercnn}
Yifan Xu, Tianqi Fan, Mingye Xu, Long Zeng, and Yu Qiao.
\newblock Spidercnn: Deep learning on point sets with parameterized convolutional filters.
\newblock In \emph{Proceedings of the European conference on computer vision (ECCV)}, pages 87--102, 2018.

\bibitem[Xue et~al.(2023)Xue, Gao, Xing, Mart{\'\i}n-Mart{\'\i}n, Wu, Xiong, Xu, Niebles, and Savarese]{xue2023ulip}
Le Xue, Mingfei Gao, Chen Xing, Roberto Mart{\'\i}n-Mart{\'\i}n, Jiajun Wu, Caiming Xiong, Ran Xu, Juan~Carlos Niebles, and Silvio Savarese.
\newblock Ulip: Learning a unified representation of language, images, and point clouds for 3d understanding.
\newblock In \emph{Proceedings of the IEEE/CVF conference on computer vision and pattern recognition}, pages 1179--1189, 2023.

\bibitem[Xue et~al.(2024)Xue, Yu, Zhang, Panagopoulou, Li, Mart{\'\i}n-Mart{\'\i}n, Wu, Xiong, Xu, Niebles, et~al.]{xue2024ulip}
Le Xue, Ning Yu, Shu Zhang, Artemis Panagopoulou, Junnan Li, Roberto Mart{\'\i}n-Mart{\'\i}n, Jiajun Wu, Caiming Xiong, Ran Xu, Juan~Carlos Niebles, et~al.
\newblock Ulip-2: Towards scalable multimodal pre-training for 3d understanding.
\newblock In \emph{Proceedings of the IEEE/CVF Conference on Computer Vision and Pattern Recognition}, pages 27091--27101, 2024.

\bibitem[Yang et~al.(2019)Yang, Wang, Clark, Hu, Wang, Markham, and Trigoni]{yang2019learning}
Bo Yang, Jianan Wang, Ronald Clark, Qingyong Hu, Sen Wang, Andrew Markham, and Niki Trigoni.
\newblock Learning object bounding boxes for 3d instance segmentation on point clouds.
\newblock \emph{Advances in Neural Information Processing Systems}, 32, 2019.

\bibitem[Yang et~al.(2023{\natexlab{a}})Yang, Wu, He, Zhao, and Liu]{yang2023sam3d}
Yunhan Yang, Xiaoyang Wu, Tong He, Hengshuang Zhao, and Xihui Liu.
\newblock Sam3d: Segment anything in 3d scenes.
\newblock \emph{arXiv preprint arXiv:2306.03908}, 2023{\natexlab{a}}.

\bibitem[Yang et~al.(2023{\natexlab{b}})Yang, Guo, Xiong, Liu, Pan, Wang, Tong, and Guo]{yang2023swin3d}
Yu-Qi Yang, Yu-Xiao Guo, Jian-Yu Xiong, Yang Liu, Hao Pan, Peng-Shuai Wang, Xin Tong, and Baining Guo.
\newblock Swin3d: A pretrained transformer backbone for 3d indoor scene understanding.
\newblock \emph{arXiv preprint arXiv:2304.06906}, 2023{\natexlab{b}}.

\bibitem[Yeshwanth et~al.(2023)Yeshwanth, Liu, Nie{\ss}ner, and Dai]{yeshwanth2023scannet++}
Chandan Yeshwanth, Yueh-Cheng Liu, Matthias Nie{\ss}ner, and Angela Dai.
\newblock Scannet++: A high-fidelity dataset of 3d indoor scenes.
\newblock In \emph{Proceedings of the IEEE/CVF International Conference on Computer Vision}, pages 12--22, 2023.

\bibitem[Yilmaz et~al.(2024)Yilmaz, Peng, Engelmann, Pollefeys, and Blum]{yilmaz2024opendas}
Gonca Yilmaz, Songyou Peng, Francis Engelmann, Marc Pollefeys, and Hermann Blum.
\newblock Opendas: Domain adaptation for open-vocabulary segmentation.
\newblock \emph{arXiv preprint arXiv:2405.20141}, 2024.

\bibitem[Yin et~al.(2023)Yin, Fu, Yang, and Lin]{yin2023or}
Youtan Yin, Zhoujie Fu, Fan Yang, and Guosheng Lin.
\newblock Or-nerf: Object removing from 3d scenes guided by multiview segmentation with neural radiance fields.
\newblock \emph{arXiv preprint arXiv:2305.10503}, 2023.

\bibitem[Yin et~al.(2024)Yin, Liu, Xiao, Cohen-Or, Huang, and Chen]{yin2024sai3d}
Yingda Yin, Yuzheng Liu, Yang Xiao, Daniel Cohen-Or, Jingwei Huang, and Baoquan Chen.
\newblock Sai3d: Segment any instance in 3d scenes.
\newblock In \emph{Proceedings of the IEEE/CVF Conference on Computer Vision and Pattern Recognition}, pages 3292--3302, 2024.

\bibitem[Zhang et~al.(2023{\natexlab{a}})Zhang, Han, Qiao, Kim, Bae, Lee, and Hong]{zhang2023faster}
Chaoning Zhang, Dongshen Han, Yu Qiao, Jung~Uk Kim, Sung-Ho Bae, Seungkyu Lee, and Choong~Seon Hong.
\newblock Faster segment anything: Towards lightweight sam for mobile applications.
\newblock \emph{arXiv preprint arXiv:2306.14289}, 2023{\natexlab{a}}.

\bibitem[Zhang et~al.(2022)Zhang, Guo, Zhang, Li, Miao, Cui, Qiao, Gao, and Li]{zhang2022pointclip}
Renrui Zhang, Ziyu Guo, Wei Zhang, Kunchang Li, Xupeng Miao, Bin Cui, Yu Qiao, Peng Gao, and Hongsheng Li.
\newblock Pointclip: Point cloud understanding by clip.
\newblock In \emph{Proceedings of the IEEE/CVF Conference on Computer Vision and Pattern Recognition}, pages 8552--8562, 2022.

\bibitem[Zhang et~al.(2023{\natexlab{b}})Zhang, Jiang, Guo, Yan, Pan, Dong, Gao, and Li]{zhang2023personalize}
Renrui Zhang, Zhengkai Jiang, Ziyu Guo, Shilin Yan, Junting Pan, Hao Dong, Peng Gao, and Hongsheng Li.
\newblock Personalize segment anything model with one shot.
\newblock \emph{arXiv preprint arXiv:2305.03048}, 2023{\natexlab{b}}.

\bibitem[Zhang et~al.(2023{\natexlab{c}})Zhang, Yang, Wang, and Li]{zhang2023growsp}
Zihui Zhang, Bo Yang, Bing Wang, and Bo Li.
\newblock Growsp: Unsupervised semantic segmentation of 3d point clouds.
\newblock In \emph{Proceedings of the IEEE/CVF Conference on Computer Vision and Pattern Recognition}, pages 17619--17629, 2023{\natexlab{c}}.

\bibitem[Zhao et~al.(2023)Zhao, Ding, An, Du, Yu, Li, Tang, and Wang]{zhao2023fast}
Xu Zhao, Wenchao Ding, Yongqi An, Yinglong Du, Tao Yu, Min Li, Ming Tang, and Jinqiao Wang.
\newblock Fast segment anything.
\newblock \emph{arXiv preprint arXiv:2306.12156}, 2023.

\bibitem[Zhou et~al.(2024)Zhou, Gu, Chiang, Xiang, and Su]{zhou2024point}
Yuchen Zhou, Jiayuan Gu, Tung~Yen Chiang, Fanbo Xiang, and Hao Su.
\newblock Point-sam: Promptable 3d segmentation model for point clouds.
\newblock \emph{arXiv preprint arXiv:2406.17741}, 2024.

\bibitem[Zhu et~al.(2023)Zhu, Zhang, He, Guo, Zeng, Qin, Zhang, and Gao]{zhu2023pointclip}
Xiangyang Zhu, Renrui Zhang, Bowei He, Ziyu Guo, Ziyao Zeng, Zipeng Qin, Shanghang Zhang, and Peng Gao.
\newblock Pointclip v2: Prompting clip and gpt for powerful 3d open-world learning.
\newblock In \emph{Proceedings of the IEEE/CVF International Conference on Computer Vision}, pages 2639--2650, 2023.

\bibitem[Zou et~al.(2024)Zou, Yang, Zhang, Li, Li, Wang, Wang, Gao, and Lee]{zou2024segment}
Xueyan Zou, Jianwei Yang, Hao Zhang, Feng Li, Linjie Li, Jianfeng Wang, Lijuan Wang, Jianfeng Gao, and Yong~Jae Lee.
\newblock Segment everything everywhere all at once.
\newblock \emph{Advances in Neural Information Processing Systems}, 36, 2024.

\end{thebibliography}
